\documentclass[10pt,twocolumn,letterpaper]{article}

\usepackage[pagenumbers]{iccv} 

\usepackage{amssymb}
\usepackage{amsmath}
\usepackage{booktabs}
\usepackage{pifont}
\usepackage{xcolor}
\usepackage{colortbl}
\usepackage{multirow}
\usepackage{subcaption}

\usepackage{algorithm}
\usepackage{algorithmicx}
\usepackage{algpseudocode}
\usepackage{amssymb}
\usepackage{amsmath}
\usepackage{amsthm}
\usepackage{booktabs}
\usepackage{pifont}
\usepackage{xcolor}
\usepackage{colortbl}
\usepackage{multirow}
\usepackage{subcaption}

\newcommand{\cmark}{\ding{51}}
\newcommand{\xmark}{\ding{55}}

\definecolor{iccvblue}{rgb}{0.21,0.49,0.74}
\usepackage[pagebackref,breaklinks,colorlinks,allcolors=iccvblue]{hyperref}
\usepackage{multirow}

\def\paperID{10886} 
\def\confName{ICCV}
\def\confYear{2025}

\title{Reg3D: Reconstructive Geometry Instruction Tuning for 3D Scene Understanding}

\author{Hongpei Zheng, Lintao Xiang, Qijun Yang, Qian Lin, Hujun Yin\\
Department of Electrical and Electronic Engineering,\\
The University of Manchester
}

\begin{document}
\maketitle
\begin{abstract}
The rapid development of Large Multimodal Models (LMMs) has led to remarkable progress in 2D visual understanding; however, extending these capabilities to 3D scene understanding remains a significant challenge. Existing approaches predominantly rely on text-only supervision, which fails to provide the geometric constraints required for learning robust 3D spatial representations. In this paper, we introduce Reg3D, a novel Reconstructive Geometry Instruction Tuning framework that addresses this limitation by incorporating geometry-aware supervision directly into the training process. Our key insight is that effective 3D understanding necessitates reconstructing underlying geometric structures rather than merely describing them. Unlike existing methods that inject 3D information solely at the input level, Reg3D adopts a dual-supervision paradigm that leverages 3D geometric information both as input and as explicit learning targets. Specifically, we design complementary object-level and frame-level reconstruction tasks within a dual-encoder architecture, enforcing geometric consistency to encourage the development of spatial reasoning capabilities. Extensive experiments on ScanQA, Scan2Cap, ScanRefer, and SQA3D demonstrate that Reg3D delivers substantial performance improvements, establishing a new training paradigm for spatially aware multimodal models.
\end{abstract}    
\section{Introduction}
The rapid advancement of large multimodal models (LMMs) has revolutionized how machines perceive and understand visual content, enabling unprecedented capabilities in handling image \cite{liBLIP2BootstrappingLanguageImage2023b,daiInstructBLIPGeneralpurposeVisionLanguage2023,zhuMiniGPT4EnhancingVisionLanguage2023,caffagniWikiLLaVAHierarchicalRetrievalAugmented2024} and video \cite{liLLaVANeXTInterleaveTacklingMultiimage2024,chengVideoLLaMA2Advancing2024,linVideoLLaVALearningUnited2024,baiQwen25VLTechnicalReport2025} tasks. However, extending these capabilities to 3D scene understanding remains a significant challenge due to the fundamental gap between 2D visual representations and complex 3D spatial relationships that require depth perception, occlusion reasoning, and multi-view geometric consistency.

Current methods for 3D scene understanding with LMMs primarily focus on input-level modifications, such as incorporating point cloud encoders \cite{xuPointLLMEmpoweringLarge2024,qiGPT4PointUnifiedFramework2025} or injecting 3D coordinates \cite{zhengVideo3DLLMLearning2025a,wangRoss3DReconstructiveVisual2025b} into multi-view images. However, we argue that these input-level enhancements alone are fundamentally insufficient for developing genuine 3D spatial awareness. The core issue lies in the inherent 2D inductive bias embedded within LMMs, which stems from their extensive pre-training on 2D visual data. This bias creates a fundamental obstacle that prevents effective integration and understanding of 3D geometric information, even when such information is explicitly provided as input.

To address this fundamental limitation, we propose \underline{\textbf{Re}}constructive \underline{\textbf{G}}eometry Instruction Tuning for \underline{\textbf{3D}} Scene Understanding (\textbf{Reg3D}), a novel training paradigm that goes beyond input modifications by incorporating 3D-aware learning objectives directly into the training process. Our key insight is that models must be explicitly guided to understand spatial relationships and comprehensive 3D layouts through dedicated pretext tasks that force them to reason about geometric structures.

Specifically, \textbf{Reg3D} introduces a 3D geometry encoder that leverages prior knowledge from pre-trained 3D foundation models, coupled with two complementary reconstruction tasks designed to bridge the gap between 2D visual understanding and 3D spatial reasoning:
(1) \textbf{Object-level 3D geometry reconstruction} requires the model to infer masked geometric information of objects across different viewpoints, fostering cross-view spatial reasoning essential for tasks like dense captioning and 3D visual grounding. (2) \textbf{Frame-level 3D geometry reconstruction} challenges the model to reconstruct depth information for entire scenes, developing comprehensive spatial understanding crucial for room layout estimation and distance reasoning.

Extensive experiments on benchmark datasets including ScanQA \cite{azumaScanQA3DQuestion2022}, Scan2Cap \cite{chenScan2CapContextawareDense2021}, and SQA3D \cite{maSQA3DSituatedQuestion2023} demonstrate that \textbf{Reg3D} achieves significant performance improvements across various 3D scene understanding tasks and establishing a new paradigm for training spatially-aware multimodal models.

In summary, our main contributions are as follows:
\begin{itemize}
    \item We propose Reconstructive Geometry Instruction Tuning (\textbf{Reg3D}), a novel training paradigm that incorporates 3D-aware learning objectives to overcome the 2D bias through dedicated geometric reconstruction tasks.
    \item We design two complementary reconstruction tasks that force models to reason about 3D spatial relationships: object-level reconstruction for cross-view spatial reasoning and frame-level reconstruction for comprehensive scene understanding.
    \item Through extensive experiments on multiple benchmark datasets, we demonstrate that \textbf{Reg3D} achieves the state-of-the-art performance across various 3D scene understanding tasks.
\end{itemize}

\section{Related Work}
\subsection{Multimodal Large Language Models(MLLMs)}
Multimodal Large Language Models (MLLMs) have shown remarkable capabilities in handling various modalities, including text, image, and video. Early works like BLIP \cite{liBLIPBootstrappingLanguageImage2022} pioneered the integration of vision and language understanding through bootstrapped pre-training with web-scale noisy data, Flamingo \cite{alayracFlamingoVisualLanguage2022} further advanced the field by introducing a novel architecture that freezes large pre-trained vision and language models while incorporating learnable cross-attention mechanisms. Subsequent models like BLIP-2 \cite{liBLIP2BootstrappingLanguageImage2023b} introduced the Q-Former architecture to bridge the modality gap more efficiently, while LLaVA \cite{liuVisualInstructionTuning2023} directly projecting visual features from a CLIP \cite{radfordLearningTransferableVisual2021} visual encoder into the language model's embedding space through linear layers. More recently, models like Qwen2.5-VL \cite{baiQwen25VLTechnicalReport2025} and LLaVA-1.5 \cite{liuImprovedBaselinesVisual2024} have demonstrated that this simple projector-based approach can achieve competitive performance while maintaining architectural simplicity. However, current MLLMs still face significant limitations in understanding and reasoning about 3D spatial scenarios. While these models excel at 2D image analysis and description, they struggle with tasks requiring depth perception, spatial relationship reasoning, and 3D scene understanding.

\subsection{3D Scene Understanding with LMMs}
The integration of 3D understanding capabilities into large language models has emerged as a critical research direction. Point-based approaches like PointLLM \cite{xuPointLLMEmpoweringLarge2024} and GPT4Point \cite{qiGPT4PointUnifiedFramework2025} directly encode 3D point clouds using specialized encoders, but suffer from the scarcity of large-scale 3D-text paired datasets compared to their 2D counterparts. Multi-view based methods such as Video3D-LLM \cite{zhengVideo3DLLMLearning2025a}, 3D-LLaVA \cite{deng3DLLaVAGeneralist3D2025}, and GPT4Scene \cite{qiGPT4SceneUnderstand3D2025} attempt to infer 3D information from multiple 2D viewpoints, but often lack explicit geometric constraints and struggle with cross-view consistency. While some approaches inject 3D coordinates directly into input sequences \cite{zhengVideo3DLLMLearning2025a,zhuLLaVA3DSimpleEffective2025} or utilize rendered bird's-eye-view representations \cite{qiGPT4SceneUnderstand3D2025}, they remain heavily dependent on high-quality 3D reconstruction data and extensive scene-text supervision, limiting their practical applicability.

To overcome these limitations, our method introduces a novel reconstructive geometry instruction tuning approach that leverages 3D geometric information not only as input but also as supervisory signals, enabling more effective learning of 3D spatial understanding without requiring massive 3D-text paired datasets.

\section{Preliminary}

A typical LLM models the canonical causal distribution of a text sequence $x = \{x_i\}_{i=1}^T$ as $p_\theta(x) = \prod_{i=1}^T p_\theta(x_i|x_{<i})$, where $\theta$ denotes the parameters and $T$ represents the sequence length. To make LLMs utilize the visual information, a common approach is to encode visual signals $v$ into a sequence of prefix tokens through an encoder like CLIP \cite{radfordLearningTransferableVisual2021} and a projector. Assume we have a $\phi$-parameterized vision encoder $\mathcal{H}_{\phi}$ and a $\psi$-parameterized projector $\mathcal{P}_{\psi}$, the visual tokens can be denoted as $v = \mathcal{H}_{\phi} \circ \mathcal{P}_{\psi}(I)$. Therefore, the canonical causal distribution of a multimodal sequence can be denoted as $p_\Theta(x) = \prod_{i=1}^T p_\Theta(x_i|x_{<i}, v)$, where $\Theta = \{\phi, \psi, \theta\}$ is the model parameters and $v \in \mathbb{R}^{L \times D}$ is the visual tokens. $L$ is the number of visual tokens and $D$ denotes the dimension of the multi-modal token.

In a typical LLM supervised fine-tuning (SFT) process, human-annotated instruction-response pairs are used as training data. Specifically, given an instruction-response text sequence $\mathcal{D} = \{x_i\}_{i=1}^T$, where $T$ is the text sequence length. The training objective is to minimize the following negative log-likelihood loss:

\begin{equation}
\mathcal{L}_{text}(\Theta=\{\phi, \psi, \theta\}) = -\frac{1}{T} \sum_{i=N+1}^{N+T} \log p_\Theta(x_i|x_{<i}, v)
\label{eq:text-loss}
\end{equation}

where $N$ is the number of visual tokens.(Assuming the visual tokens are prepended to the sequence.) From Eq. \ref{eq:text-loss}, we can see that only the text tokens after the visual tokens ($x_{i>N}$) are used for supervised training.
\section{Preliminary}

A typical LLM models the canonical causal distribution of a text sequence $x = \{x_i\}_{i=1}^T$ as $p_\theta(x) = \prod_{i=1}^T p_\theta(x_i|x_{<i})$, where $\theta$ denotes the parameters and $T$ represents the sequence length. To make LLMs utilize the visual information, a common approach is to encode visual signals $v$ into a sequence of prefix tokens through an encoder like CLIP \cite{radfordLearningTransferableVisual2021} and a projector. Assume we have a $\phi$-parameterized vision encoder $\mathcal{H}_{\phi}$ and a $\psi$-parameterized projector $\mathcal{P}_{\psi}$, the visual tokens can be denoted as $v = \mathcal{H}_{\phi} \circ \mathcal{P}_{\psi}(I)$. Therefore, the canonical causal distribution of a multimodal sequence can be denoted as $p_\Theta(x) = \prod_{i=1}^T p_\Theta(x_i|x_{<i}, v)$, where $\Theta = \{\phi, \psi, \theta\}$ is the model parameters and $v \in \mathbb{R}^{L \times D}$ is the visual tokens. $L$ is the number of visual tokens and $D$ denotes the dimension of the multi-modal token.

In a typical LLM supervised fine-tuning (SFT) process, human-annotated instruction-response pairs are used as training data. Specifically, given an instruction-response text sequence $\mathcal{D} = \{x_i\}_{i=1}^T$, where $T$ is the text sequence length. The training objective is to minimize the following negative log-likelihood loss:

\begin{equation}
\mathcal{L}_{text}(\Theta=\{\phi, \psi, \theta\}) = -\frac{1}{T} \sum_{i=N+1}^{N+T} \log p_\Theta(x_i|x_{<i}, v)
\label{eq:text-loss}
\end{equation}

where $N$ is the number of visual tokens.(Assuming the visual tokens are prepended to the sequence.) From Eq. \ref{eq:text-loss}, we can see that only the text tokens after the visual tokens ($x_{i>N}$) are used for supervised training.
\section{Methodology}

In this paper, we propose a novel method for indoor scene understanding, named \textbf{Reg3D}. In contrast to the existing 2D LMMs solely injecting 3D geometry information and supervised by output text, \textbf{Reg3D} using geometry information as the reconstruction target for supervised training. In this section, we first introduce the overall framework of \textbf{Reg3D}, followed by detailed descriptions of its key components.

\begin{figure*}[!t]
    \centering
    \includegraphics[width=0.9\linewidth]{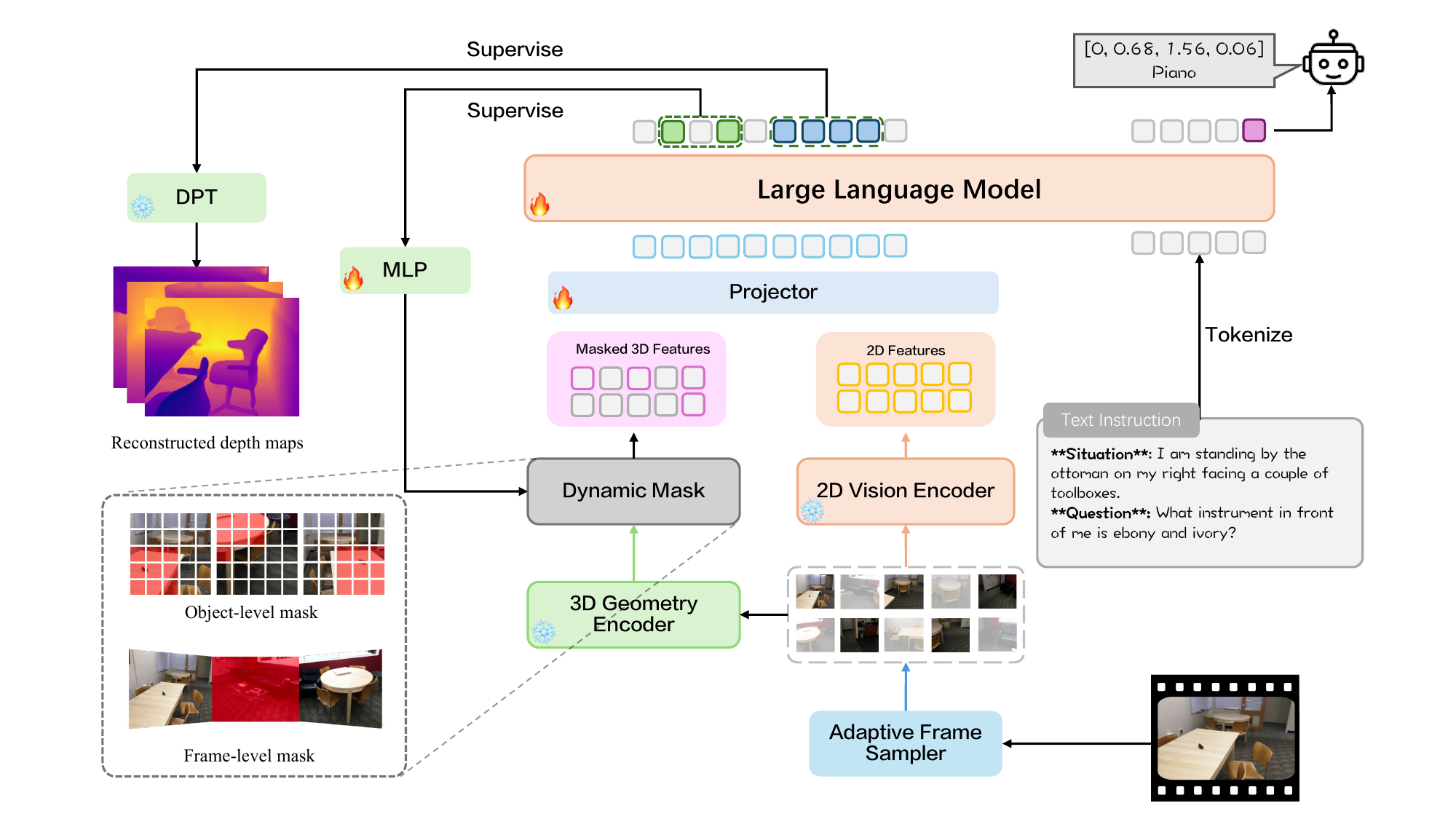}
    \caption{Overview of the \textbf{Reg3D} framework. By requiring the model to reconstruct object features in regions occluded by object-level masks and depth maps in frames occluded by frame-level masks, we explicitly guide the model to focus on and learn 3D geometric representations.}
    \label{fig:overview}
\end{figure*}

\subsection{Overview}
As illustrated in Fig.\ref{fig:overview}, the overall framework of \textbf{Reg3D} consists of five main components: a 2D vision encoder $\mathcal{E}_{\phi}$, a 3D geometry encoder $\mathcal{H}_{\theta}$, a large language model $\mathcal{L}_{\tau}$, a projector $P_{\psi}$, and a DPT module $\mathcal{D}_{\zeta}$.

The key innovation of \textbf{Reg3D} lies in its use of 3D geometric information as both input and supervisory signals. Unlike existing methods that only use text outputs $x_{i>N}$ as training targets, \textbf{Reg3D} introduces reconstruction tasks that explicitly guide the model to learn 3D spatial relationships. Specifically, during training, we mask portions of the 3D geometric features and require the model to reconstruct them based on available 2D visual features and unmasked 3D geometric information. The reconstruction loss can be formulated as:

\begin{equation}
\mathcal{L}_{recon}(x, I; \Theta) = \mathcal{D}(\mathcal{L}_{\tau}(\mathcal{E}_{\phi}(I)+\overline{\mathcal{M}} \circ \mathcal{H}_{\theta}(I)), \mathcal{M} \circ \mathcal{H}_{\theta}(I))
\label{eq:recon-loss}
\end{equation}

where $\mathcal{D}$ and $\mathcal{M}$ is the distance function and the mask.

During the reconstruction task, LMMs are required to recover the masked geometry feature map $\mathcal{M} \circ \mathcal{H}_{\theta}(I)$ based on the given 2D features $\mathcal{E}_{\phi}(I)$ and the unmasked geometry features $\overline{\mathcal{M}} \circ \mathcal{H}_{\theta}(I)$. This differ from \cite{huangMLLMsNeed3DAware2025}, which directly aligning the LLMs' output $x_{i \le N}$ with the geometry features

In the following sections, we detail the dual-encoder architecture and introduce two complementary reconstruction tasks that enable effective 3D geometry learning.

\subsection{Dual-encoder Architecture} \label{sec:dual-encoder-architecture}

To enable VLMs trained on 2D image data to perceive 3D geometric information, we propose a dual-encoder architecture that consists of a 2D vision encoder $\mathcal{E}_{\phi}$ and a 3D geometry encoder $\mathcal{H}_{\theta}$. 

As shown in Fig. \ref{fig:overview}, the 2D vision encoder  follows the standard vision transformer (ViT) architecture \cite{dosovitskiyImageWorth16x162021}, which divides the input image into non-overlapping patches and processes them through multiple transformer layers. We adopt the same architecture as the Qwen2.5-VL \cite{baiQwen25VLTechnicalReport2025}, but unlike Qwen2.5-VL which concatenates 2 adjacent video frames, we process each frame as an individual image:

\begin{equation}
\mathcal{F}_{2D} = \mathcal{E}_{\phi}(\{I_i\}_{i=1}^N), \quad \mathcal{F}_{2D} \in \mathbb{R}^{N \times \frac{H}{p_{2D}} \times \frac{W}{p_{2D}} \times D_{2D}}
\label{eq:2d-feature}
\end{equation}

where $p_{2D}$ denotes the 2D encoder patch size and $D_{2D}$ represents the dimension of the visual tokens.

3D geometry encoder $\mathcal{H}_{\theta}$ is designed to extract rich geometric representations from video frames. Specifically, we utilize VGGT \cite{wangVGGTVisualGeometry2025}, a state-of-the-art 3D reconstruction model, which employs a transformer-based architecture to simultaneously infer multiple 3D attributes including camera parameters, point maps, and depth maps. To obtain spatially-aligned geometric features, we extract the output from VGGT's final transformer layer, which encodes comprehensive 3D scene geometry in a patchified format:

\begin{equation}
\mathcal{F}_{3D} = \mathcal{H}_{\theta}(\{I_i\}_{i=1}^N), \quad \mathcal{F}_{3D} \in \mathbb{R}^{N \times \frac{H}{p_{3D}} \times \frac{W}{p_{3D}} \times D_{3D}}
\label{eq:3d-feature}
\end{equation}

where $p_{3D}$ denotes the 3D encoder patch size and $D_{3D}$ represents the dimension of the 3D geometry tokens.

The outputs from both encoders are then combined through a fusion module to align the visual and geometric features. Qwen2.5-VL uses a two-layer multi-layer perceptron (MLP) to concatenate adjacent sets of four patches, which means $p_{2D}$ is 2 times larger than $p_{3D}$. We follow the same strategy to compress the 3D geometry feature:

\begin{equation}
\mathcal{F}'_{3D} = P_{\psi}(\mathcal{F}_{3D}), \quad \mathcal{F}'_{3D} \in \mathbb{R}^{N \times \frac{H}{2 \times p_{3D}} \times \frac{W}{2 \times p_{3D}} \times D_{3D}'}
\label{eq:3d-feature-fusion}
\end{equation}

\begin{figure*}[t]
    \centering
    \includegraphics[width=\textwidth, trim=0cm 2.5cm 0cm 0cm, clip]{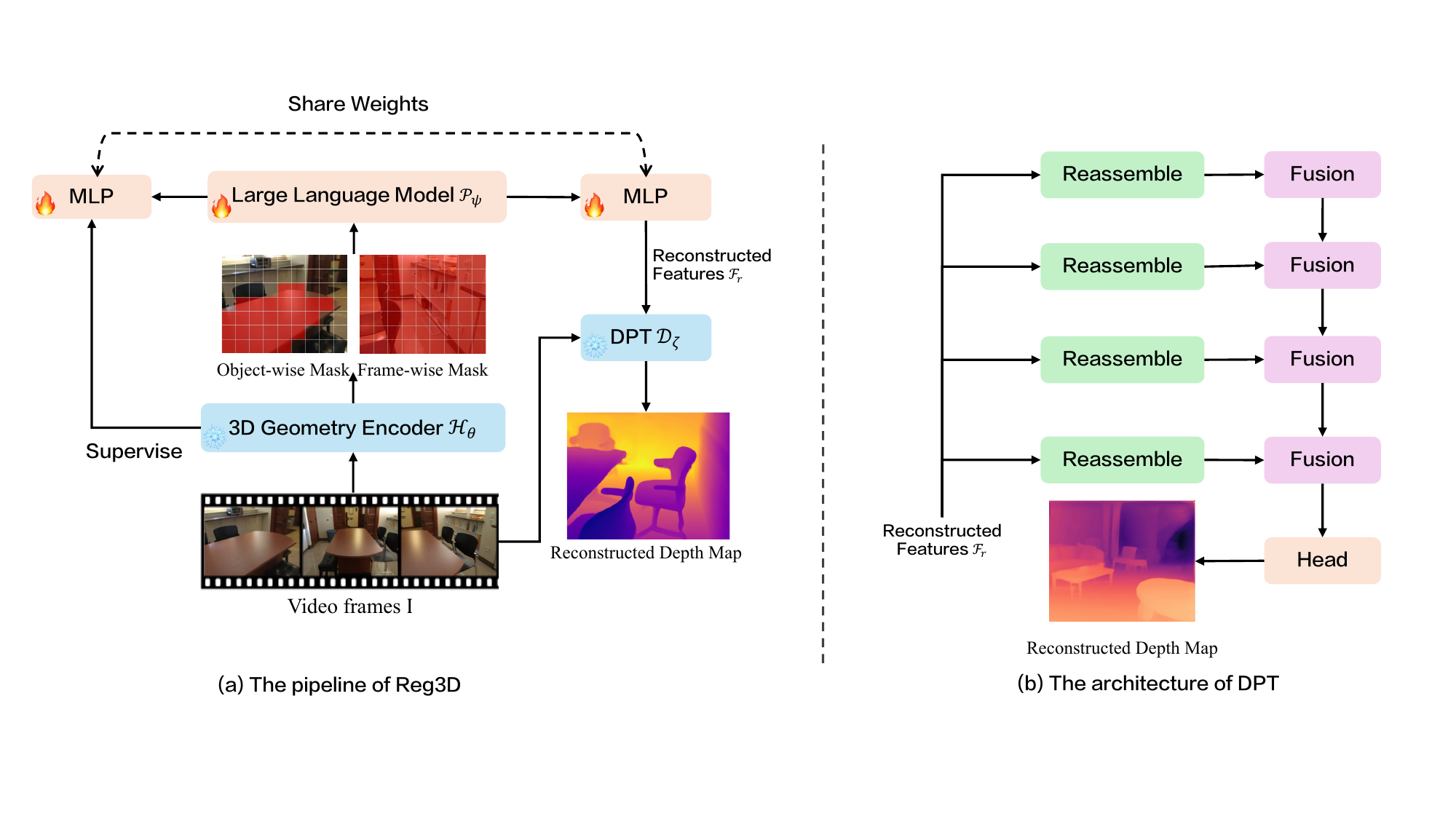}
    \caption{(a) \textbf{Reg3D} pipeline: The 3D geometry encoder extracts geometric features from video frames, which are partially masked and fused with 2D features by the large language model to enable reconstructive supervision. (b) DPT architecture: The reconstructed features $\mathcal{F}_{r}$ is processed by multiple Reassemble and Fusion modules, followed by a Head to generate the reconstructed depth map.}
    \label{fig:mask_pipeline}
\end{figure*}

\subsection{Object-level Reconstruction} \label{sec:object-level-reconstruction}

Object-level reconstruction enforces cross-view geometric consistency by learning to reconstruct masked 3D features of object instances across multiple viewpoints. This task is grounded in the fundamental principle of multi-view geometry: corresponding 3D points should maintain consistent geometric properties regardless of the observing camera pose, thereby enabling the model to develop view-invariant spatial representations.

\noindent \textbf{Object Selection and Masking Strategy.} We first identify salient objects in each scene using preprocessed 2D segmentation results, filtering out background such as walls and floors. During training, we randomly select a subset of objects for reconstruction. For each selected object that appears in $K$ views ($K \leq N$), we employ a partial masking strategy: retain 3D geometric features in $M$ randomly chosen views ($M < K$) while masking the 3D features in the remaining $(K-M)$ views. Specifically, we only mask the patches that spatially overlap with the selected objects based on their 2D segmentation masks, ensuring that the reconstruction task focuses on learning object-specific geometric representations rather than arbitrary scene regions.

\noindent \textbf{Patch-based Formulation.} We divide each frame into non-overlapping patches of size $p \times p$. Let $g_i$ denote the $i$-th patch across all $N$ views, resulting in a total of $N \times \frac{H}{p} \times \frac{W}{p}$ patches:

\begin{equation}
\{g_i\}_{i=1}^{N \times \frac{H}{p} \times \frac{W}{p}}, \quad g_i \in \mathbb{R}^{p^2 \times 3}
\label{eq:object-level-reconstruction-input}
\end{equation}

We generate a binary mask $\mathcal{M} \in \{0, 1\}^{N \times \frac{H}{p} \times \frac{W}{p}}$ based on the object masking strategy, where 0 indicates masked patches and 1 indicates unmasked patches. The multimodal tokens are then constructed as:

\begin{equation}
v_i = \begin{cases}
    \mathcal{E}_{\phi}(g_i) + \mathcal{P}_{\psi} \circ \mathcal{H}_{\theta}(g_i), & \text{if } \mathcal{M}_i = 1 \\
    \mathcal{E}_{\phi}(g_i), & \text{if } \mathcal{M}_i = 0
\end{cases}
\label{eq:object-level-reconstruction-mask}
\end{equation}

\noindent \textbf{Reconstruction Objective.} For masked patches, the model must reconstruct the missing 3D geometric features $\mathcal{P}_{\psi} \circ \mathcal{H}_{\theta}(g_i)$ using available 2D visual information and unmasked 3D context. Let $G$ denote the total number of masked patches and $v_i^{re}$ is the reconstructed features generated by the language model. Our training objective minimizes:

\begin{equation}
\mathcal{L}_{object} = \frac{\alpha}{G} \sum_{i: \mathcal{M}_i=0} \mathcal{D}(v_i^{target}, v_i^{re})
\label{eq:object-level-reconstruction-loss}
\end{equation}

where $v_i^{target} = \mathcal{E}_{\phi}(g_i) + \mathcal{P}_{\psi} \circ \mathcal{H}_{\theta}(g_i)$ represents the ground truth complete multimodal token, $\alpha$ is a weighting hyperparameter, and $\mathcal{D}$ is the distance function defined as:

\begin{equation}
\mathcal{D}(v_i^{target}, v_i^{re}) = -\frac{v_i^{target} \cdot v_i^{re}}{||v_i^{target}|| \cdot ||v_i^{re}||}
\label{eq:object-level-reconstruction-distance}
\end{equation}

This cosine similarity-based loss encourages the model to generate reconstructed features that are semantically aligned with the ground truth geometric representations.

Detailed algorithm procedures for object-level reconstruction are provided in the appendix.

\subsection{Frame-level Reconstruction} \label{sec:frame-level-reconstruction}

Frame-level reconstruction complements object-level reconstruction by operating at the view level rather than the patch level. This task masks the 3D geometric features of entire views and reconstructs them using information from other views, enabling the model to learn holistic spatial relationships and develop the ability to infer 3D geometric structure from 2D visual cues.

\noindent \textbf{Depth-based Reconstruction Framework.} Unlike object-level reconstruction that focuses on local geometric consistency, frame-level reconstruction leverages depth estimation as a proxy for 3D geometric understanding. We employ a Depth Prediction Transformer (DPT) module adapted from VGGT \cite{wangVGGTVisualGeometry2025} to generate scene depth maps. The DPT module takes as input the multimodal representations generated by the language model and produces depth estimates for the corresponding views.

\noindent \textbf{Masking and Reconstruction Process.} As shown in Figure \ref{fig:mask_pipeline},  given a set of $N$ input views $\mathcal{I} = \{I_i\}_{i=1}^N$, we randomly select $K$ views ($K < N$) for masking. For these masked views, we remove their 3D geometric features from the multimodal tokens, retaining only the 2D visual features:

\begin{equation}
v_i^{masked} = \mathcal{E}_{\phi}(g_i), \quad \text{for masked views } i \in \mathcal{M}_{frame}
\label{eq:frame-level-masking}
\end{equation}

where $\mathcal{M}_{frame}$ denotes the set of masked view indices.

The reconstruction process utilizes the complete multimodal tokens from unmasked views $\{v_j\}_{j \notin \mathcal{M}_{frame}}$ along with the 2D features from masked views to predict depth maps. Let $D_i^{gt} \in \mathbb{R}^{H \times W}$ denote the ground truth depth map for view $i$, and $\hat{D}_i$ represent the reconstructed depth map generated by the DPT module:

\begin{equation}
\hat{D}_i = \text{DPT}(\{v_j\}_{j \notin \mathcal{M}_{frame}}, v_i^{masked}), \quad \text{for } i \in \mathcal{M}_{frame}
\label{eq:frame-level-reconstruction}
\end{equation}

\noindent \textbf{Training Objective.} The frame-level reconstruction loss is formulated as:

\begin{equation}
\mathcal{L}_{frame} = \frac{\beta}{K} \sum_{i \in \mathcal{M}_{frame}} ||D_i^{gt} - \hat{D}_i||_2^2
\label{eq:frame-level-reconstruction-loss}
\end{equation}

where $\beta$ is a weighting hyperparameter and $||\cdot||_2^2$ denotes the squared L2 norm. This L2 loss encourages pixel-level accuracy in depth prediction, which is crucial for maintaining geometric fidelity in 3D scene understanding.

\subsection{Overall training objective}

Our complete training framework integrates reconstructive geometry learning with standard language modeling through a multi-task objective. The total loss function combines three complementary components:

\begin{equation}
\mathcal{L}_{total} = \mathcal{L}_{text} + \lambda_1 \mathcal{L}_{object} + \lambda_2 \mathcal{L}_{frame}
\label{eq:total-loss}
\end{equation}

Here, $\lambda_1$ and $\lambda_2$ denote hyperparameters that regulate the relative importance of the individual loss components."

This approach allows the model to effectively transfer the prior knowledge of 3D spatial understanding from the pre-trained 3D geometry encoder while preserving the strong foundation in contextual reasoning and 2D visual understanding from the base model.

\subsection{Adaptive Frame Sampling}

Due to GPU memory constraints, \textbf{Reg3D} processes 8-32 frames per scene, while ScanNet \cite{daiScanNetRichlyAnnotated3D2017} scenes contain over 1500 frames on average. Unlike uniform sampling that selects frames at fixed intervals, we propose an adaptive sampling strategy that maximizes spatial coverage.

Given a video sequence with $N$ frames, we aim to select $K$ frames ($K \ll N$) that provide optimal scene coverage. Our approach consists of three steps: (1) We first uniformly sample $M$ candidate frames ($K < M \ll N$) from the sequence. (2) For each candidate frame, we extract 3D point clouds using VGGT \cite{wangVGGTVisualGeometry2025} and merge them into a unified scene representation. (3) Unlike previous methods \cite{hu3DLLMMemLongTermSpatialTemporal2025,zhengVideo3DLLMLearning2025a} that directly compute voxel coverage, we project the merged point cloud onto each camera plane using z-buffering to handle occlusion relationships. This transforms frame selection into a maximum coverage problem, solved greedily by iteratively selecting frames that cover the most uncovered projected points.

This z-buffer-based projection approach ensures that selected frames account for realistic visibility constraints and collectively capture comprehensive scene geometry with diverse viewpoints. More details are provided in the appendix.

\begin{table*}[t]
\centering
\footnotesize  
\setlength{\tabcolsep}{4pt} 

\begin{tabular}{l|c|ccccc|cc}
\toprule
Methods & Video & \multicolumn{5}{c|}{ScanQA (val)} & \multicolumn{2}{c}{SQA3D (test)} \\
& Input & BLEU-4 & METEOR & ROUGE-L & CIDEr & EM & EM & EM-R \\
\midrule
\multicolumn{9}{l}{\textit{Expert Models}} \\
ScanQA & \xmark & 10.1 & 13.1 & 33.3 & 64.9 & 21.1 & 47.2 & - \\
SQA3D & \xmark & 11.2 & 13.5 & 34.5 & - & - & 46.6 & - \\
3D-Vista & \xmark & - & - & - & 69.6 & 22.4 & 48.5 & - \\
\midrule
\multicolumn{9}{l}{\textit{3D/2.5D-Input Models}} \\
3D-LLM & \xmark & 12.0 & 14.5 & 35.7 & 69.4 & 20.5 & - & - \\
LL3DA & \xmark & 13.5 & 15.9 & 37.3 & 76.8 & - & - & - \\
Chat-Scene & \xmark & 14.3 & 18.0 & 41.6 & 87.7 & 21.6 & 54.6 & 57.5 \\
3D-LLaVA & \xmark & 17.1 & 18.4 & 43.1 & 92.6 & - & 54.5 & 56.6 \\
Video-3D LLM & \xmark & 16.2 & 19.8 & 49.0 & 102.1 & 30.1 & 58.6 & - \\
\midrule
\multicolumn{9}{l}{\textit{Video-Input Models}} \\
Qwen2.5-VL-72B & \cmark & 12.0 & 13.0 & 35.2 & 66.9 & - & 47.0 & 50.9 \\
Oryx-34B & \cmark & - & 15.0 & 37.3 & 72.3 & - & - & - \\
GPT4Scene-HDM & \cmark & 15.5 & 18.9 & 46.5 & 96.3 & - & 59.4 & 62.4 \\
\rowcolor{gray!20} \textbf{Reg3D} & \cmark & \textbf{18.3} & \textbf{20.2} & \textbf{49.1} & \textbf{104.7} & \textbf{30.3} & \textbf{60.0} & \textbf{62.9} \\
\bottomrule
\end{tabular}
\caption{\textbf{Evaluation of 3D question answering} on ScanQA, SQA3D. \textit{Expert models} refer to task-specific models. \textit{3D/2.5D-input models} leverage 3D priors, such as point clouds and depth maps, for inference, whereas \textit{video-input models} rely solely on video frames as input.
\label{tab:performance_comparison}}
\end{table*}
\section{Experiments}
\subsection{Datasets}
We evaluate the performance of \textbf{Reg3D} on 4 benchmarks that covers main 3D scene understanding tasks: ScanQA \cite{azumaScanQA3DQuestion2022} focuses on answering questions about 3D indoor scenes with complex reasoning requirements; ScanRefer \cite{chenScanRefer3DObject2020} targets 3D object localization based on natural language descriptions in indoor environments; SQA3D \cite{azumaScanQA3DQuestion2022} emphasizes situated question answering that requires understanding both visual and spatial context in 3D scenes; Scan2Cap \cite{chenScan2CapContextawareDense2021} focus on dense caption tasks. All datasets are sourced from ScanNet \cite{daiScanNetRichlyAnnotated3D2017}, which is a large-scale 3D indoor scene dataset containing 1513 scenes with 2.5 million frames.

\subsection{Evaluation Metrics}
For ScanQA \cite{azumaScanQA3DQuestion2022}, we report CIDEr \cite{vedantamCIDErConsensusbasedImage2015}, BLEU \cite{papineniBleuMethodAutomatic2002}, exact match (EM) accuracy, METEOR \cite{banerjeeMETEORAutomaticMetric2005} and ROUGE-L \cite{linROUGEPackageAutomatic2004} scores. For ScanRefer \cite{chenScanRefer3DObject2020}, we report the accuracy at IoU thresholds of 0.25 and 0.5 (IoU@0.25 and IoU@0.5). Scan2Cap \cite{chenScan2CapContextawareDense2021} is evaluated by CIDEr and BLEU-4 scores at IoU@0.5 (C@0.5, B-4@0.5). For SQA3D \cite{azumaScanQA3DQuestion2022}, we report EM accuracy and refined exact match (EM-R) accuracy.

\subsection{Implementation Details}
We build our \textbf{Reg3D} based on Qwen2.5-VL-7B \cite{baiQwen25VLTechnicalReport2025}, we utilize the Qwen2.5-VL-7B's vision module as our 2D vision encoder. We use VGGT \cite{wangVGGTVisualGeometry2025} as 3D geometry encoder, both the 2D and 3D branches are freezed during training. The model is fine-tuned on combination dataset of ScanQA \cite{azumaScanQA3DQuestion2022}, ScanRefer \cite{chenScanRefer3DObject2020}, SQA3D \cite{azumaScanQA3DQuestion2022}, Multi3DRefer \cite{zhangMulti3DReferGroundingText2023} for 1 epoch. The model is optimized using AdamW \cite{loshchilovDecoupledWeightDecay2019} optimizer with a batch size of 48. The learning rate is peak at 1e-5 and weight decay is set to 0.01. All experiments are conducted with 6$\times$A100-80G. The frame resolution is set to 504$\times$392. We use ground truth object bounding boxes as training target, while in inference, we employ Mask3D \cite{schultMask3DMaskTransformer2023} to refine our 3D visual grounding task.

\subsection{Comparison with State-of-the-Art Methods}
Table \ref{tab:performance_comparison} and \ref{tab:dense_caption} present comprehensive comparisons of \textbf{Reg3D} with state-of-the-art methods across multiple 3D scene understanding tasks. We categorize baseline methods into three groups: \textit{Expert models} are task-specific architectures including ScanQA \cite{azumaScanQA3DQuestion2022}, SQA3D \cite{maSQA3DSituatedQuestion2023}, MVT \cite{chenMVTMultiviewVision2021}, 3DJCG \cite{cai3DJCGUnifiedFramework2022}, M3DRef-CLIP \cite{zhangMulti3DReferGroundingText2023a} and 3D-Vista \cite{zhu3DVisTAPretrainedTransformer2023}. \textit{3D/2.5D-Input Models} require 3D priors such as point clouds or depth maps, including 3D-LLM \cite{hong3DLLMInjecting3D2023}, LL3DA \cite{chenLL3DAVisualInteractive2024}, Chat-Scene \cite{huangChatSceneBridging3D2024}, 3D-LLaVA \cite{deng3DLLaVAGeneralist3D2025} and Video-3D LLM \cite{zhengVideo3DLLMLearning2025a}. \textit{Video-Input Models} utilize only RGB video frames, including Qwen2.5-VL-72B \cite{baiQwen25VLTechnicalReport2025}, GPT4Scene-HDM \cite{qiGPT4SceneUnderstand3D2025} and Oryx-34B \cite{liuOryxMLLMOnDemand2025}.

\noindent \textbf{3D Question Answering.} On ScanQA validation set (Table \ref{tab:performance_comparison}), \textbf{Reg3D} achieves competitive performance with scores of 18.3 BLEU-4, 20.2 METEOR, 49.1 ROUGE-L, 104.7 CIDEr, and 30.3 EM. Notably, our method outperforms most 3D/2.5D-input models while using only video frames as input, demonstrating superior practicality. On SQA3D test set, \textbf{Reg3D} achieves 60.0 EM and 62.9 EM-R, showing strong performance in situated question answering tasks.

\begin{table}[h]
\centering
\small
\begin{tabular}{l|cc|cc}
\toprule
Methods & \multicolumn{2}{c|}{ScanRefer} & \multicolumn{2}{c}{Scan2Cap} \\
\cmidrule{2-5}
& A@0.25 & A@0.5 & B-4@0.5 & C@0.5 \\
\midrule
\textit{Expert Models} & & & & \\
ScanRefer & 37.3 & 24.3 & - & - \\
MVT & 40.8 & 33.3 & - & - \\
Scan2Cap & - & - & 22.4 & 35.2 \\
ViL3DRel & 47.9 & 37.7 & - & - \\
3DJCG & 49.6 & 37.3 & 31.0 & 49.5 \\
M3DRef-CLIP & 51.9 & 38.4 & - & - \\
\midrule
\textit{LMMs} & & & & \\
3D-LLM & 30.3 & - & - & - \\
Ground 3D-LLM & 47.9 & 44.1 & 35.5 & 70.6 \\
ChatScene & 55.5 & 50.2 & 36.3 & 77.1 \\
LLaVA-3D & 54.1 & 42.4 & 41.1 & 79.2 \\
Video-3D-LLM & \textbf{58.1} & \textbf{51.7} & 41.3 & 83.8 \\
\rowcolor{gray!20} \textbf{Reg3D} & 53.8 & 47.3 & \textbf{43.1} & \textbf{87.9} \\
\bottomrule
\end{tabular}
\caption{\textbf{Evaluation of 3D visual grounding} on Scan2Cap and ScanRefer.\label{tab:dense_caption}}
\end{table}

\noindent \textbf{3D Dense Captioning.} Table \ref{tab:dense_caption} shows results on Scan2Cap dataset. \textbf{Reg3D} achieves 43.1 BLEU-4 and 87.9 CIDEr, outperforming the previous best method Video-3D-LLM by 1.8 BLEU-4 and 4.1 CIDEr. This demonstrates our model's ability to generate high-quality descriptions for 3D objects in complex indoor scenes.

\begin{figure}[h]
    \centering
    \includegraphics[width=\linewidth,trim={8cm 0cm, 9cm, 0cm},clip]{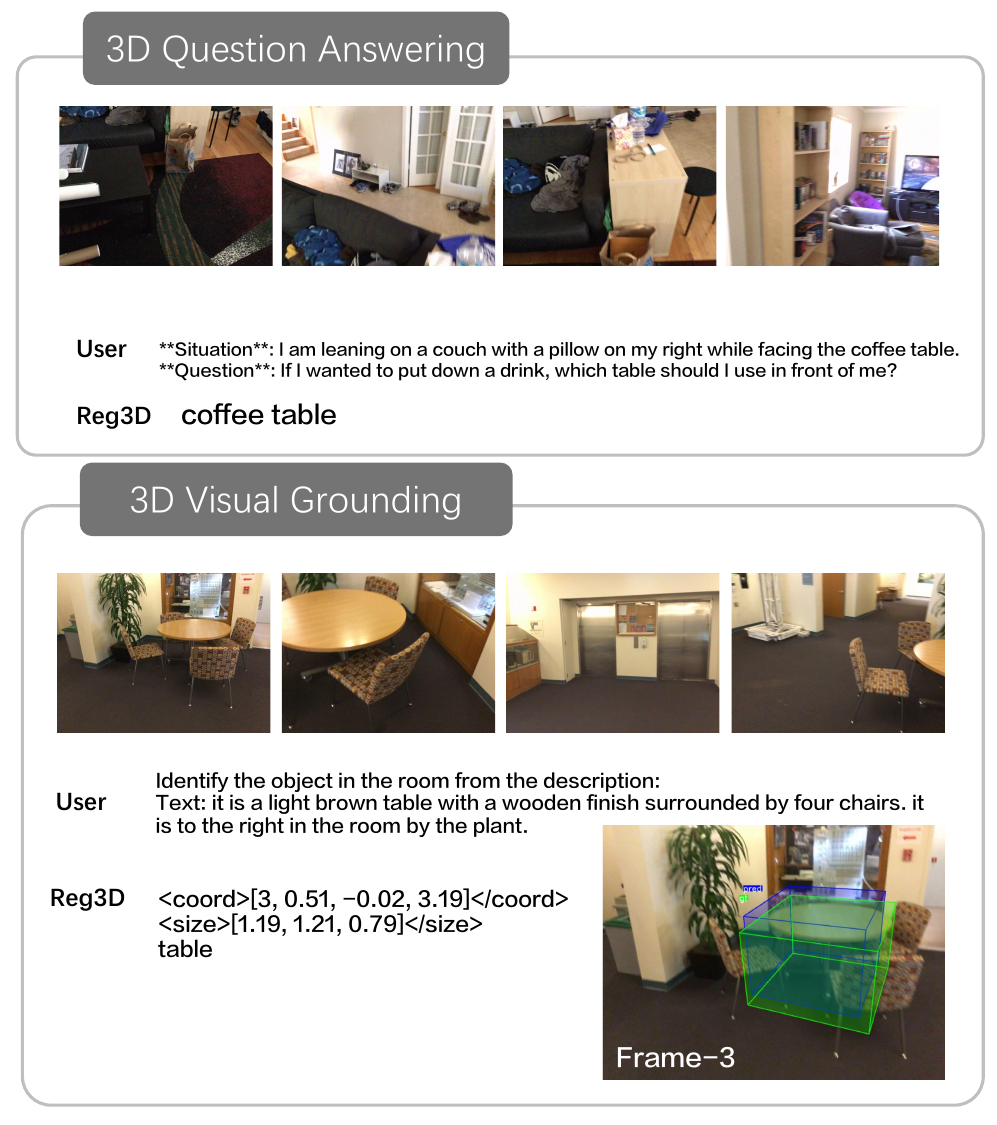}
    \caption{Visual grounding example. Given a natural language description, our model can accurately locate the corresponding 3D object. The blue and green bounding boxes indicate the predicted 3D object and the ground truth 3D object respectively.}
    \label{fig:visual_grounding}  
    \end{figure}

\noindent \textbf{3D Visual Grounding.} On ScanRefer dataset (Table \ref{tab:dense_caption}), \textbf{Reg3D} achieves 53.8 Acc@0.25 and 47.3 Acc@0.5. While competitive, the performance is slightly lower than some methods. This is primarily due to our end-to-end design that directly outputs bounding box coordinates in text format (Figure \ref{fig:visual_grounding}), whereas other methods typically generate 3D proposals first and then perform matching.


\subsection{Ablation Study}

\noindent \textbf{Effectiveness of 3D geometry information.} In Table \ref{tab:ablation_geometry}, we demonstrate that incorporating 3D geometry information significantly improves model performance. Specifically, adding the 3D geometry encoder brings consistent gains across all metrics, with notable improvements of +4.2 CIDEr and +1.8 EM scores on ScanQA, as well as +2.5 EM and +2.2 EM-R scores on SQA3D. These substantial improvements validate the importance of integrating pretrained 3D foundation model into our framework.

\begin{table}[h]
\centering
\small
\begin{tabular}{l|ccc|cc}
\toprule
\multirow{2}{*}{Method} & \multicolumn{3}{c|}{ScanQA} & \multicolumn{2}{c}{SQA3D} \\
& CIDEr & B-4 & EM & EM & EM-R \\
\midrule
w/o Geo. Enc. & 97.4 & 14.9 & 28.3 & 56.7 & 59.9 \\
w/ Geo. Enc. & \textbf{101.6} & \textbf{16.6} & \textbf{30.1} & \textbf{59.2} & \textbf{62.1} \\
\midrule
$\Delta$ & +4.2 & +1.7 & +1.8 & +2.5 & +2.2 \\
\bottomrule
\end{tabular}
\caption{Ablation study on the effectiveness of 3D geometry encoder. Results are reported on ScanQA and SQA3D validation sets. All training is conducted with same dataset and same training parameters. Reconstructive geometry instruction tuning is not applied.\label{tab:ablation_geometry}}
\end{table}

\begin{table}[h]
\centering

\small
\begin{tabular}{l|cc|cc}
\toprule
\multirow{2}{*}{Method} & \multicolumn{2}{c|}{ScanQA} & \multicolumn{2}{c}{SQA3D} \\
& CIDEr & EM & EM & EM-R \\
\midrule
$L_{text}$ & 101.6 & 30.1 & 59.2 & 62.1 \\
$L_{frame} + L_{text}$ & 102.2 & 30.9 & 58.9 & 61.8 \\
$L_{obj} + L_{text}$ & 103.5 & 30.2 & \textbf{60.7} & \textbf{63.0} \\
$L_{frame} + L_{obj} + L_{text}$ & \textbf{104.7} & \textbf{30.3} & 60.0 & 62.9 \\
\bottomrule
\end{tabular}
\caption{Ablation study on the effectiveness of reconstruction supervision. Results are reported on ScanQA and SQA3D. \label{tab:ablation_recon}} 
\end{table}

\noindent \textbf{Effectiveness of Reconstruction Supervision.} Table~\ref{tab:ablation_recon} presents the ablation study on the effectiveness of reconstruction supervision. The table compares four different training settings: using only the text loss ($L_{text}$), adding frame-level reconstruction loss ($L_{frame} + L_{text}$), adding object-level reconstruction loss ($L_{obj} + L_{text}$), and combining both frame-level and object-level reconstruction losses ($L_{frame} + L_{obj} + L_{text}$). Results are reported on the ScanQA and SQA3D.

From the results, we observe that introducing either frame-level or object-level reconstruction supervision leads to performance improvements over the baseline that uses only text loss. Notably, the combination of both reconstruction losses achieves the best overall performance. For example, on ScanQA, the CIDEr score increases from 101.6 to 104.7, and the EM score improves from 30.1 to 30.3. On SQA3D, both EM and EM-R metrics also show consistent gains. These results demonstrate that reconstruction supervision effectively enhances the model's spatial understanding and contributes to better performance on downstream 3D scene understanding tasks.
\section{Conclusion}


In this paper, we proposed \textbf{Reg3D}, a novel Reconstructive Geometry Instruction Tuning framework for 3D scene understanding. Our approach combines a dual-encoder architecture with complementary object-level and frame-level reconstruction tasks that leverage 3D geometric information as both input and supervisory signals. Through extensive experiments on benchmark datasets including ScanQA, Scan2Cap, ScanRefer, and SQA3D, we demonstrate that \textbf{Reg3D} achieves significant performance improvements across various 3D scene understanding tasks while using only RGB video frames as input. The ablation studies validate the effectiveness of our 3D geometry encoder and reconstruction supervision components, showing that both object-level and frame-level reconstruction tasks contribute meaningfully to the overall performance gains.

Our work establishes a new paradigm for training spatially-aware multimodal models, demonstrating that reconstructive geometry instruction tuning can effectively bridge the gap between 2D visual understanding and 3D spatial reasoning. This framework opens up promising directions for future research in 3D scene understanding and can potentially be extended to other 3D understanding tasks and modalities.

{
    \small
    \bibliographystyle{ieeenat_fullname}
    \bibliography{main}
}

\clearpage
\setcounter{page}{1}
\maketitlesupplementary

\section{More Implementation Details}

\subsection{Adaptive Frame Sampler}

Given a video stream $I_m=\{I_i\}_{i=1}^M$, our goal is to sample $I_k=\{I_i\}_{i=1}^K$ frames that maximize the scene information coverage. We first uniformly sample $N$ frames from the video. Then we perform 3D reconstruction on these frames using VGGT \cite{wangVGGTVisualGeometry2025}, obtaining camera parameters $g_i \in \mathbb{R}^9$ (intrinsic and extrinsic parameters) and point maps $P_i \in \mathbb{R}^{3 \times H \times W}$ through VGGT's camera head and point head. We then merge the point maps from all frames to obtain a scene point map $P \in \mathbb{R}^{3 \times M \times H \times W}$. To accelerate computation, we divide the space into voxel grids, where points within the same voxel are merged. After merging, we project the point cloud onto a 2D plane. For a 3D point $P=(X,Y,Z)$, its projection onto the image plane can be computed as:

\begin{equation}
    \begin{bmatrix} x \\ y \\ 1 \end{bmatrix} = C_i[R|t] \begin{bmatrix} X \\ Y \\ Z \\ 1 \end{bmatrix}
\end{equation}

where $C_i$ is the camera intrinsic matrix and $[R|t]$ is the camera extrinsic matrix. Due to the sparsity of the point cloud, we treat each projected point as a square with side length $d$. For a projected point $p=(x,y)$, its coverage area can be expressed as:

\begin{equation}
    A(p) = \{(u,v) | |u-x| \leq \frac{d}{2}, |v-y| \leq \frac{d}{2}\}
\end{equation}

To handle occlusions between points at different depths, we employ a z-buffer technique where points closer to the camera take precedence over points that are further away. Specifically, for each pixel location $(u,v)$ in the projected image, we maintain a depth value $z(u,v)$ recording the distance of the closest point that projects to that location. When a new point projects to $(u,v)$, it only updates the coverage if its depth is smaller than the current $z(u,v)$.

For each view $i$, we can obtain a set of visible points $V_i$ that are covered by this view:

\begin{equation}
    V_i = \{p_j | p_j \text{ is visible in view } i\}
\end{equation}

where $p_j$ represents the $j$-th point in the merged point cloud. A point is considered visible if it falls within the camera's field of view and is not occluded by other points according to the z-buffer.

With the visibility information of each view, we can formulate our frame selection problem as a maximum coverage problem: given a collection of sets $\{V_i\}_{i=1}^M$ and an integer $K$, find $K$ sets whose union has the maximum cardinality. This is a well-known NP-hard problem, but can be efficiently solved using a greedy algorithm that iteratively selects the view that covers the most uncovered points.

\begin{figure*}[t]
    \centering
    \includegraphics[trim={5cm 4.5cm 5cm 3cm}, clip, width=\linewidth]{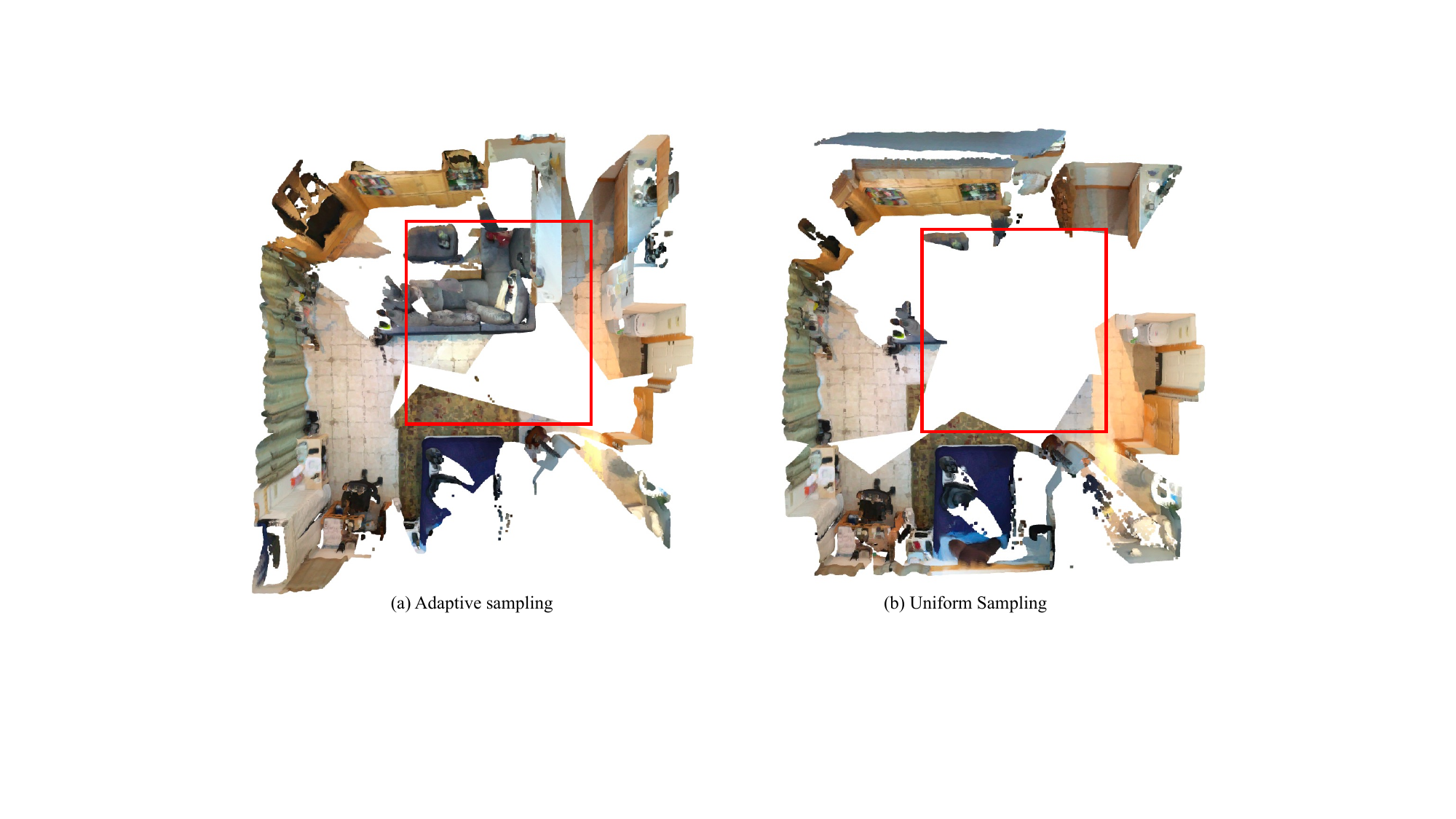}
    \caption{Visualization of adaptive frame sampling and uniform sampling. Compared to uniform sampling (b) which leads to significant information loss, adaptive frame sampling (a) achieves better coverage of scene details by selecting key frames that evenly capture different viewpoints. Results shown with $K=8$ and $N=128$.}
    \label{fig:frame_sampling}
\end{figure*}

\begin{algorithm}
\caption{Adaptive Frame Sampling}
\begin{algorithmic}[1]
\Require Video frames $\mathcal{I}=\{I_i\}_{i=1}^N$, target frame count $K$
\Ensure Selected frames $I_k=\{I_i\}_{i=1}^K$
\State Uniformly sample $M$ frames: $I_m \gets \text{UniformSample}(\mathcal{I}, M)$
\State $\{g_i, P_i\}_{i=1}^M \gets \text{VGGT}(I_m)$ \Comment{Get camera params and point maps}
\State $P \gets \text{MergePointMaps}(\{P_i\}_{i=1}^M)$ \Comment{Merge point maps}
\State $P \gets \text{VoxelGridDownsample}(P)$ \Comment{Voxel grid downsampling}
\For{$i \gets 1$ to $M$}
    \State Project points using camera $g_i$ and compute z-buffer
    \State $V_i \gets$ visible points in view $i$
\EndFor
\State $S \gets \emptyset$ \Comment{Selected frame indices}
\State $C \gets \emptyset$ \Comment{Covered points}
\For{$k \gets 1$ to $K$}
    \State $maxGain \gets 0$ \Comment{Maximum coverage gain}
    \State $bestView \gets -1$ \Comment{Best view index}
    \For{$i \gets 1$ to $M$}
        \If{$i \notin S$}
            \State $gain \gets |V_i \setminus C|$ \Comment{Count new points covered}
            \If{$gain > maxGain$}
                \State $maxGain \gets gain$
                \State $bestView \gets i$
            \EndIf
        \EndIf
    \EndFor
    \State $S \gets S \cup \{bestView\}$ \Comment{Add best view to selection}
    \State $C \gets C \cup V_{bestView}$ \Comment{Update covered points}
\EndFor
\State \Return $\{I_i\}_{i \in S}$
\end{algorithmic}
\end{algorithm}

The results of our adaptive frame sampling approach can be visualized in Figure \ref{fig:frame_sampling}. As shown in the figure, our method effectively selects keyframes that provide maximal coverage of the scene geometry, ensuring comprehensive capture of the 3D structure. The selected frames tend to be well-distributed across different viewpoints of the scene, avoiding redundant views while preserving important geometric details.

\subsection{A.2 Object-level Reconstruction}

In our experiments, we select 3 salient objects per scene for reconstruction. Specifically, we first identify salient objects using preprocessed 2D segmentation results while filtering out background elements such as walls and floors. For each selected object, we employ a partial masking strategy: given K views where the object appears, we preserve 3D geometric features in the view that contains the most image patches overlapping with the object's 2D segmentation mask, while masking features in all other views. This ensures we keep the most informative view for each object while masking features in remaining views, allowing the model to learn object-specific geometric representations.

The detailed algorithm procedure is as follows:

\begin{algorithm}[h]
\caption{Object-level Reconstruction}
\begin{algorithmic}[1]
\Require Video frames $\{I_i\}_{i=1}^N$, segmentation masks $\{M_i\}_{i=1}^N$
\Ensure Reconstructed 3D geometric features
\State Identify salient objects and filter background
\State Randomly select 3 objects for reconstruction
\For{each selected object}
    \State Identify K views containing the object
    \State Count overlapping patches in each view
    \State Select view with maximum patch overlap
    \State Mask corresponding regions in all other views
    \State Only mask patches overlapping with object mask
\EndFor
\State Use LLM to reconstruct masked 3D geometric features
\State \Return Reconstructed 3D geometric features
\end{algorithmic}
\end{algorithm}

\subsection{Evaluation Details}

\begin{figure*}[t]
    \centering
    \begin{subfigure}{\linewidth}
        \includegraphics[trim={5cm 2cm 5cm 3cm}, clip, width=\linewidth]{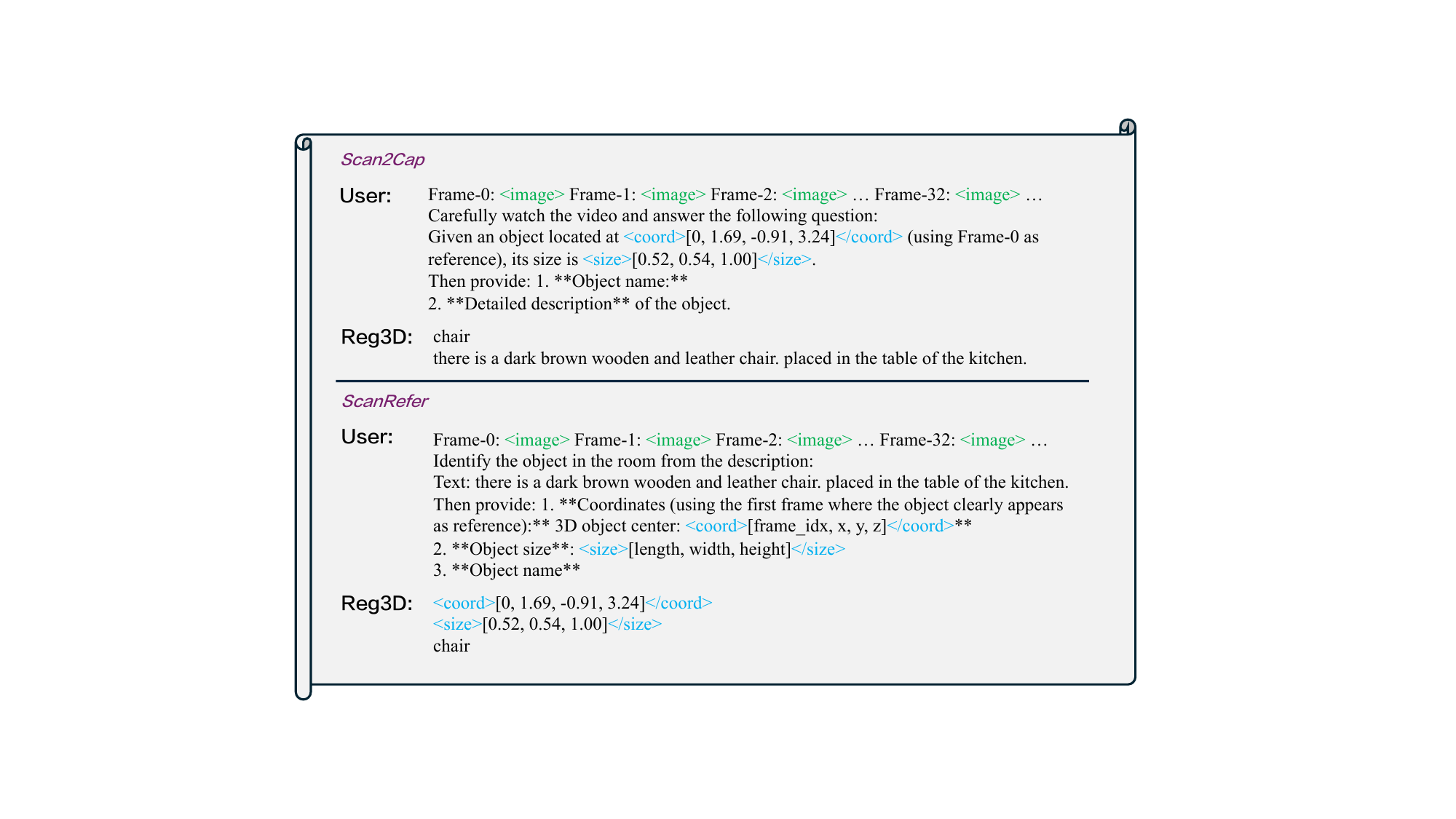}
        \caption{Prompt for \textbf{3D dense captioning} and \textbf{3D visual grounding} tasks.}
        \label{fig:3d_vg_prompt}
    \end{subfigure}
    \begin{subfigure}{\linewidth}
        \includegraphics[trim={5.5cm 4cm 5cm 3cm}, clip, width=\linewidth]{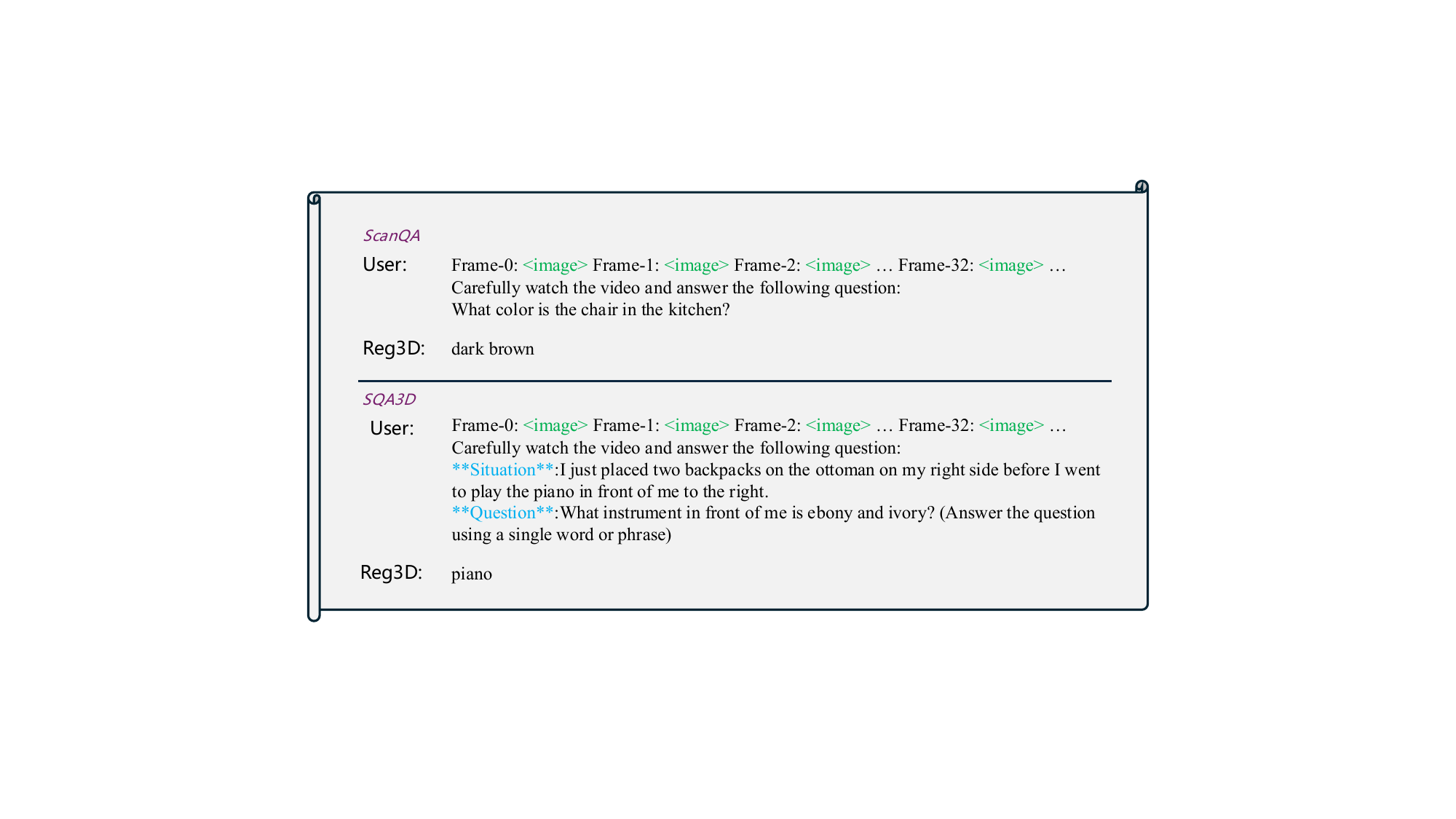}
        \caption{Prompt for \textbf{3D question answering} tasks.}
        \label{fig:3d_qa_prompt}
    \end{subfigure}
\end{figure*}

Unlike prior approaches \cite{zhuLLaVA3DSimpleEffective2025,zhengVideo3DLLMLearning2025a} that formulate 3D visual grounding as a classification task, our Reg3D directly outputs bounding box information through two pairs of special tokens \texttt{<coord></coord>} and \texttt{<size></size>} to predict the 3D coordinates and dimensions of target objects. For coordinates, we adopt camera coordinate system where coordinates are represented as \texttt{<coord>frame\_idx, x, y, z</coord>} with \texttt{frame\_idx} indicating the index of the image frame.

For \textbf{3D Visual Grounding} tasks, we convert the problem into a two-step process: first identifying the video frame where the target object is initially clearly visible, then predicting its bounding box in the image coordinate system of that frame. This approach avoids complex classification directly in 3D space, instead simplifying the task to 2D grounding while preserving the spatial information of the 3D scene. 
To further improve localization accuracy, we leverage the pre-trained Mask3D \cite{schultMask3DMaskTransformer2023} model for object detection and segmentation in the scene. Specifically, we first obtain 3D bounding boxes and segmentation masks for all objects using Mask3D. During inference, we match our model's predicted bounding boxes with Mask3D detection results and select the detection box with the highest Intersection-over-Union (IoU) as the final output. 
For cases where the IoU between our model's predicted bounding box and Mask3D detection results is below 0.25, we leverage the semantic labels output by our model to further filter Mask3D detection boxes. Specifically, we first filter Mask3D detection boxes based on semantic category matching, then select the detection box with the closest center point distance as the final output. This post-processing strategy effectively improves localization accuracy, especially in cases where target objects are partially occluded.

For \textbf{3D Dense Captioning} tasks, we follow previous works \cite{zhengVideo3DLLMLearning2025a,zhuLLaVA3DSimpleEffective2025} to decompose the task into two stages: object detection and description generation. First, we utilize Mask3D-detected object proposals to obtain object center coordinates in the reference frame's coordinate system by applying the extrinsic matrix transformation. Then, our model generates object descriptions based on these coordinates through greedy sampling. Finally, we evaluate the generated captions using the COCO caption evaluation toolkit to compute CIDEr, BLEU, Rouge, and METEOR scores.

For \textbf{3D Question Answering} tasks, we use greedy decoding to generate answers and evaluate the CIDEr, BLEU, Rouge, and METEOR scores between generated answers and ground truth answers.

The detailed prompt for each task is shown in Figure \ref{fig:3d_vg_prompt} and Figure \ref{fig:3d_qa_prompt}.

\subsection{Training Costs}
We conduct all experiments using 6 NVIDIA A100 GPUs with gradient accumulation steps of 8 and per-GPU batch size of 1. In Table \ref{tab:training_costs}, we analyze the training costs of our model with and without reconstruction supervision. The results show that adding reconstruction supervision only introduces marginal computational overhead. Specifically, the training speed increases from 35.2 seconds per iteration to 40.2 seconds per iteration, representing a 1.14× slowdown. Meanwhile, the GPU memory usage increases slightly from 74.5G to 76.7G, only a 1.03× increase. This demonstrates that our reconstruction supervision achieves significant performance improvements with reasonable additional computational costs.

\begin{table}[h]

\centering


\begin{tabular}{cccc}

\toprule

$L_{frame}$ & $L_{obj}$ & Speed (s/iter) & GPU Memory \\

\midrule

- & - & 35.2 & 74.5 G \\

\cmark & \cmark & 40.2 (1.14×) & 76.7 G (1.03×) \\ 

\bottomrule

\end{tabular}

\caption{Training costs comparison with and without reconstruction supervision. \label{tab:training_costs}}

\end{table}

\section{More Experiments}

\subsection{More Results}

\begin{table*}[t]
    \centering
    \begin{tabular}{l|cccccc|c|c}
    \toprule
    \multirow{2}{*}{Method} & \multicolumn{6}{c|}{Question Type} & \multirow{2}{*}{Avg. (EM)} & \multirow{2}{*}{EM-R} \\
    & What & Is & How & Can & Which & Others & & \\
    \midrule
    \multicolumn{9}{l}{\textbf{Expert Models}} \\
    SQA3D & 31.6 & 63.8 & 46.0 & 69.5 & 43.9 & 45.3 & 46.6 & -- \\
    3D-VisTA & 34.8 & 63.3 & 45.4 & 69.8 & 47.2 & 48.1 & 48.5 & -- \\
    \midrule
    \multicolumn{9}{l}{\textbf{2D LLMs}} \\
    InternVL2-8B & 30.5 & 53.8 & 5.5 & 47.3 & 25.8 & 36.3 & 33.0 & 45.3 \\
    Qwen2-VL-7B & 29.0 & 59.2 & 33.4 & 50.5 & 44.2 & 43.2 & 40.7 & 46.7 \\
    LLaVA-Video-7B & 42.7 & 56.3 & 47.5 & 55.3 & 50.1 & 47.2 & 48.5 & -- \\
    \midrule
    \multicolumn{9}{l}{\textbf{3D LMMs}} \\
    LEO & -- & -- & -- & -- & -- & -- & 50.0 & 52.4 \\
    Scene-LLM & 40.9 & 69.1 & 45.0 & 70.8 & 47.2 & 52.3 & 54.2 & -- \\
    ChatScene & 45.4 & 67.0 & 52.0 & 69.5 & 49.9 & 55.0 & 54.6 & 57.5 \\
    LLaVA-3D & -- & -- & -- & -- & -- & -- & 55.6 & -- \\
    Video-3D-LLM & 51.1 & 72.4 & 55.5 & \textbf{69.8} & 51.3 & 56.0 & 58.6 & -- \\
    \textbf{Reg3D} & \textbf{53.5} & \textbf{75.6} & \textbf{57.1} & 67.9 & \textbf{52.6} & \textbf{57.4} & \textbf{60.0} & \textbf{62.9} \\
    \bottomrule
    \end{tabular}
    \caption{Full results of \textbf{3D question answering} on SQA3D dataset.}
    \label{tab:3dqa_full}
\end{table*}

The detailed experimental results in Tables \ref{tab:3dqa_full} and \ref{tab:scanqa_full} demonstrate that our proposed Reg3D achieves state-of-the-art performance across most evaluation metrics. These comprehensive results validate the effectiveness of our reconstructive geometry instruction tuning approach in enhancing 3D scene understanding capabilities.

\begin{table*}[t]
    \centering
    \begin{tabular}{l|c|cccc|cccc}
    \toprule
    \multirow{2}{*}{Method} & \multirow{2}{*}{EM} & \multicolumn{4}{c|}{BLEU-n Metrics} & \multicolumn{4}{c}{Language Generation Metrics} \\
    & & BLEU-1 & BLEU-2 & BLEU-3 & BLEU-4 & ROUGE & METEOR & CIDEr \\
    \midrule
    \multicolumn{9}{l}{\textbf{Expert Models}} \\
    ScanQA & 21.1 & 30.2 & 20.4 & 15.1 & 10.1 & 33.3 & 13.1 & 64.9 \\
    3D-VLP & 21.7 & 30.5 & 21.3 & 16.7 & 11.2 & 34.5 & 13.5 & 67.0 \\
    3D-VisTA  & -- & -- & -- & -- & 13.9 & 35.7 & -- & -- \\
    \midrule
    \multicolumn{9}{l}{\textbf{2D LLMs}} \\
    InternVL2-8B & 16.9 & 20.0 & 9.8 & 5.2 & 2.7 & 32.6 & 14.5 & 55.3 \\
    Qwen2-VL-7B & 19.0 & 27.8 & 13.6 & 6.3 & 3.0 & 34.2 & 11.4 & 53.9 \\
    LLaVA-Video-7B & -- & 39.7 & 26.6 & 9.3 & 3.1 & 44.6 & 17.7 & 88.7 \\
    \midrule
    \multicolumn{9}{l}{\textbf{3D LMMs}} \\
    3D-LLM & 20.5 & 39.3 & 25.2 & 18.4 & 12.0 & 35.7 & 14.5 & 69.4 \\
    Chat-3D & -- & 29.1 & -- & -- & 6.4 & 28.5 & 11.9 & 53.2 \\
    LL3DA & -- & -- & -- & -- & 13.5 & 37.3 & 15.9 & 76.8 \\
    LEO & 24.5 & -- & -- & -- & 11.5 & 39.3 & 16.2 & 80.0 \\
    Scene-LLM & 27.2 & 43.6 & 26.8 & 19.1 & 12.0 & 40.0 & 16.6 & 80.0 \\
    ChatScene & 21.6 & 43.2 & 29.1 & 20.6 & 14.3 & 41.6 & 18.0 & 87.7 \\
    Grounded 3D-LLM & -- & -- & -- & -- & 13.4 & -- & -- & 72.7 \\
    LLaVA-3D & 27.0 & -- & -- & -- & 14.5 & 50.1 & 20.7 & 91.7 \\
    Video-3D-LLM & 30.1 & 47.1 & 31.7 & 22.8 & 16.2 & 49.0 & 19.8 & 102.1 \\
    \textbf{Reg3D} & \textbf{30.3} & \textbf{47.2} & \textbf{31.9} & \textbf{23.0} & \textbf{18.3} & \textbf{49.1} & \textbf{20.2} & \textbf{104.7} \\
    \bottomrule
    \end{tabular}
    \caption{Full results of \textbf{3D question answering} on ScanQA dataset.}
    \label{tab:scanqa_full}
\end{table*}

\section{More Visualizations}
\begin{figure*}[h]
    \centering
    \begin{subfigure}{\linewidth}
        \includegraphics[trim={0cm 3cm 0cm 3cm}, clip, width=\linewidth]{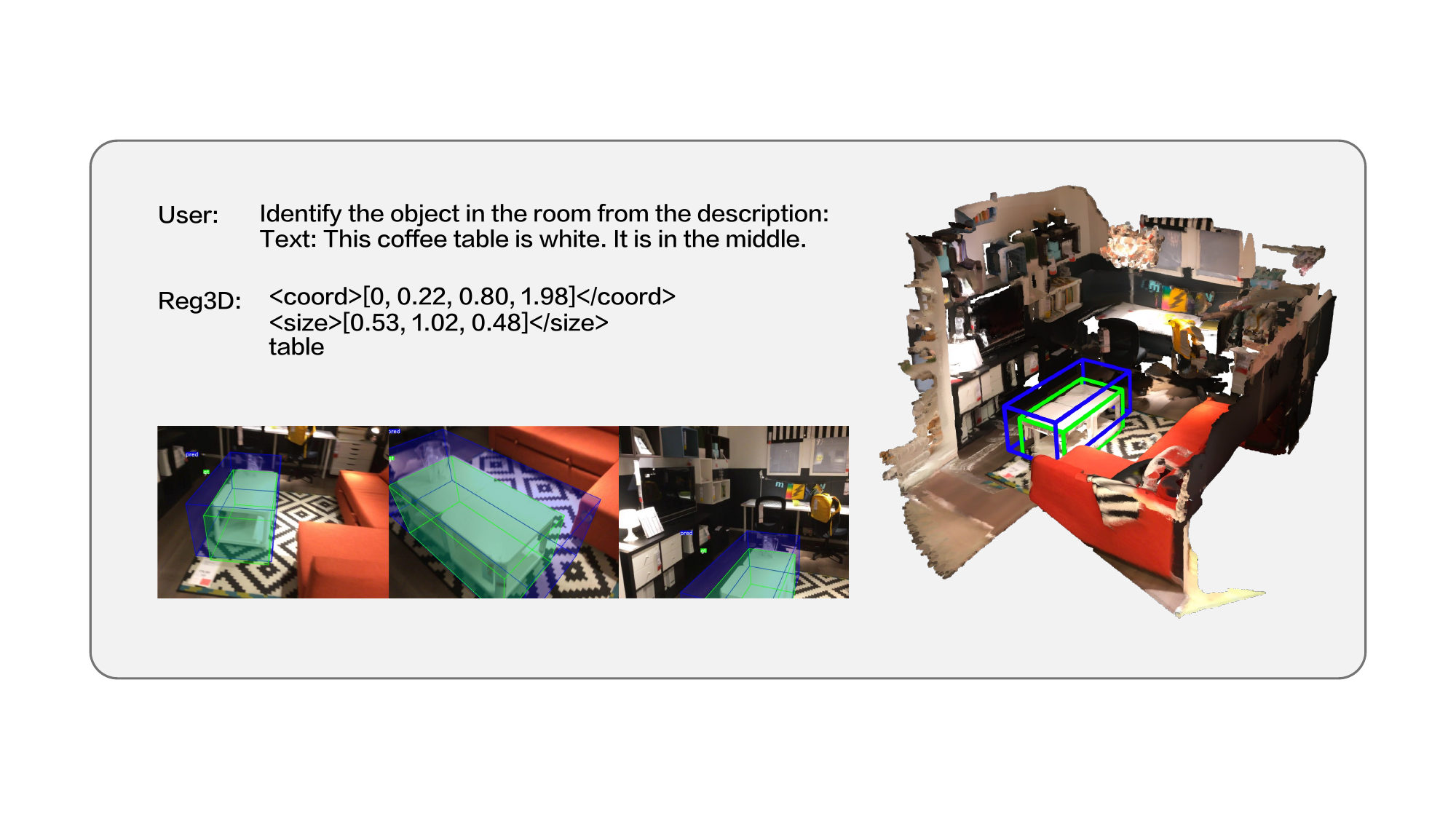}
    \end{subfigure}
    \begin{subfigure}{\linewidth}
        \includegraphics[trim={0cm 3cm 0cm 3cm}, clip, width=\linewidth]{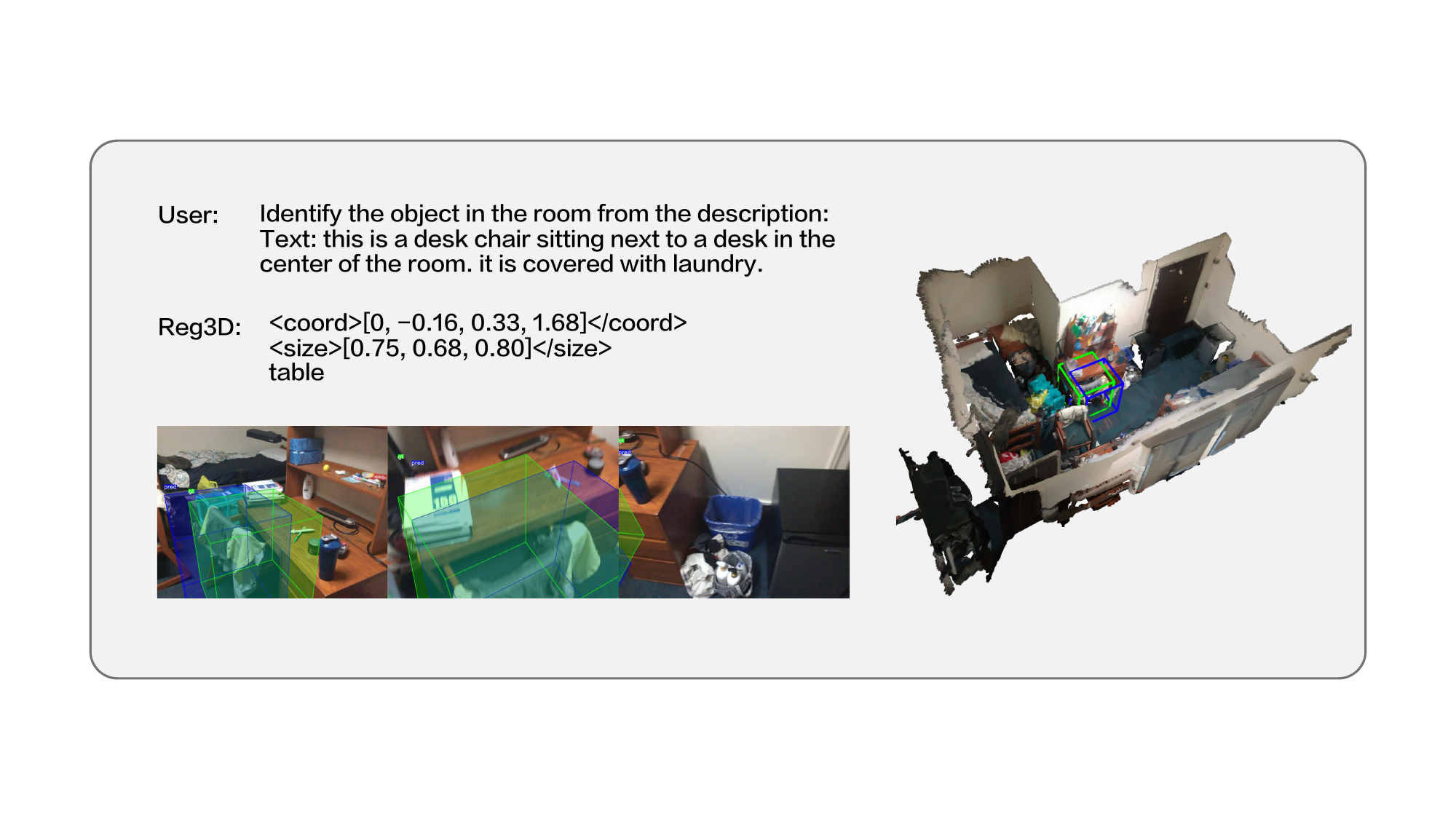} 
    \end{subfigure}
    \begin{subfigure}{\linewidth}
        \includegraphics[trim={0cm 3cm 0cm 3cm}, clip, width=\linewidth]{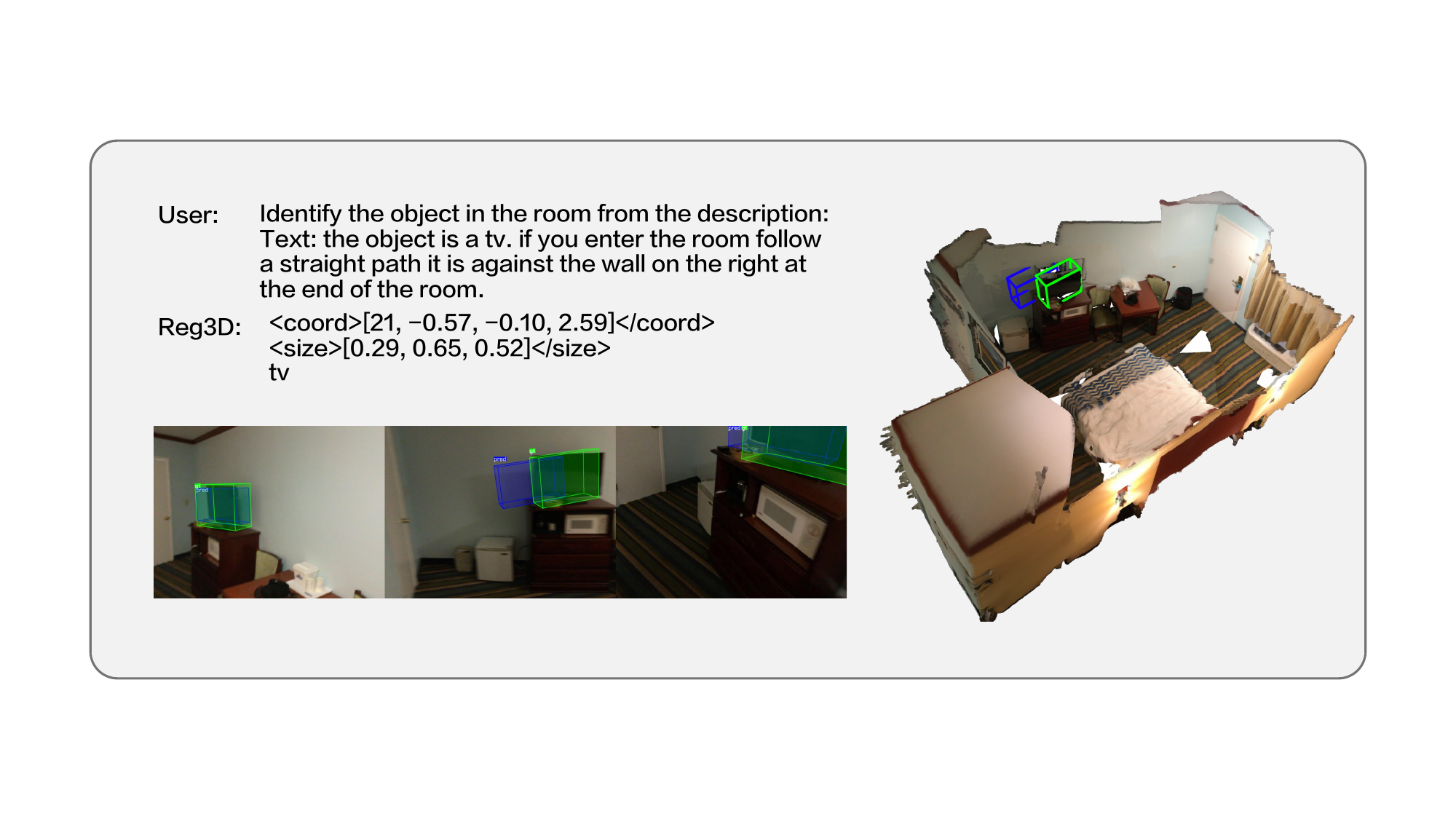}
    \end{subfigure}
    \caption{Visualization of 3D visual grounding results on ScanRefer. Blue boxes denote the predicted bounding boxes, while green boxes denote the ground truth bounding boxes. The first and second rows denote successful cases. The third row case shows the direct prediction failure case, but the model can still generate correct bounding boxes based on semantic category matching.}
    \label{fig:vg_visual}
\end{figure*}

In Figure~\ref{fig:vg_visual}, we present the visualization results of 3D visual grounding on the ScanRefer dataset. The blue boxes indicate the predicted bounding boxes by our model, while the green boxes represent the ground truth annotations. The first two rows demonstrate successful cases where our model accurately localizes the target objects. The third row shows a case where direct prediction fails, but the model can still generate correct bounding boxes through semantic category matching.

\end{document}